\documentclass[runningheads]{llncs}
\usepackage[T1]{fontenc}
\usepackage{graphicx}
\usepackage{multirow}
\usepackage{booktabs}
\usepackage{bigstrut}
\usepackage{amsmath} 

\usepackage{bm}
\usepackage{multirow}
\usepackage[table]{xcolor}
\usepackage{overpic}
\usepackage{makecell}

\usepackage{booktabs}
\usepackage{bigstrut}

\usepackage{times}
\usepackage{epsfig}
\usepackage{graphicx}
\usepackage{amsmath}
\usepackage{amssymb}
\usepackage[accsupp]{axessibility}  
\usepackage{float}
\usepackage{bm}

\usepackage{algorithmic}
\usepackage{array}
\usepackage{fixltx2e}
\usepackage{url}

\usepackage{gensymb}

\usepackage{color}
\begin{document}
\title{TDM: Temporally-Consistent Diffusion Model for All-in-One Real-World Video Restoration}
\titlerunning{TDM: Temporally-Consistent Diffusion Model}
%
\author{Yizhou Li\inst{1}\orcidID{0000-0002-7122-2087} \and
Zihua Liu\inst{1}\orcidID{0000-0002-3908-0517} \and
Yusuke Monno\inst{1}\orcidID{0000-0001-6733-3406} \and
Masatoshi Okutomi\inst{1}\orcidID{0000-0001-5787-0742}}

%
\authorrunning{Y. Li et al.}
%
\institute{Institute of Science Tokyo, Tokyo, Japan\\
\email{\{yli,zliu,ymonno\}@ok.sc.e.titech.ac.jp,\\
mxo@ctrl.titech.ac.jp}}
%
\maketitle              
\begin{abstract}
In this paper, we propose the first diffusion-based all-in-one video restoration method that utilizes the power of a pre-trained Stable Diffusion and a fine-tuned ControlNet. Our method can restore various types of video degradation with a single unified model, overcoming the limitation of standard methods that require specific models for each restoration task.
Our contributions include an efficient training strategy with Task Prompt Guidance (TPG) for diverse restoration tasks, an inference strategy that combines Denoising Diffusion Implicit Models~(DDIM) inversion with a novel Sliding Window Cross-Frame Attention (SW-CFA) mechanism for enhanced content preservation and temporal consistency, and a scalable pipeline that makes our method all-in-one to adapt to different video restoration tasks. Through extensive experiments on five video restoration tasks, we demonstrate the superiority of our method in generalization capability to real-world videos and temporal consistency preservation over existing state-of-the-art methods.
Our method advances the video restoration task by providing a unified solution that enhances video quality across multiple applications.

\end{abstract}

\section{Introduction} \label{sec: introduction}
Video restoration is crucial as videos often lose quality due to factors like adverse weather, noise, compression, and limited sensor resolution, which can significantly hamper computer vision tasks such as object detection and video surveillance. Existing video restoration methods \cite{viddenoise_task,vidsr,mfqev2,vrt,rvrt,fastdvd,rawviddenoise,semividdenoise_nturain} have shown progress but are limited to specific degradations, requiring separate models for each restoration task. This is costly and impractical for real-world applications where multiple degradations may occur. A unified all-in-one video restoration model is highly demanded for practical use.
Recent attempts to create a single model for all-in-one image restoration have been promising \cite{airnet,promptir}, but they fall short in video applications due to the challenge of maintaining temporal consistency.
The main demands for video restoration are (1)~developing a single model that can handle various degradations, (2)~ensuring temporal consistency of restored videos, and (3)~achieving robust real-world performance.


\begin{figure*}
  \includegraphics[width=\textwidth]{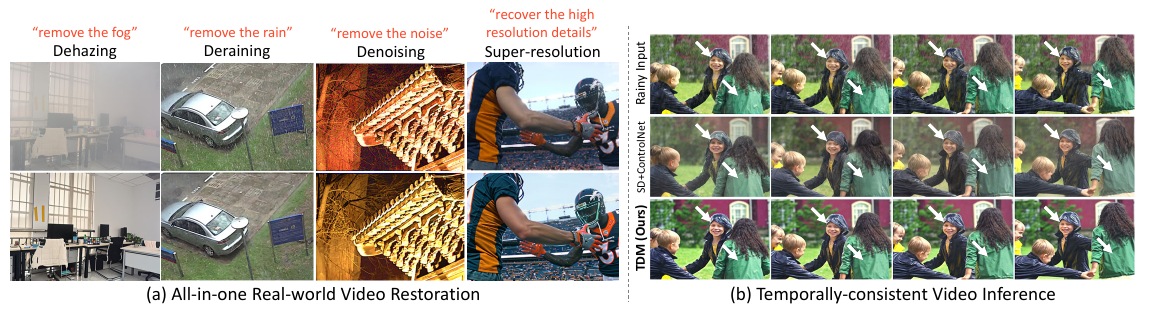}
  \vspace{-6mm}
  \caption{Our Temporally-consistent Diffusion Model~(TDM) has two main features: (a) Our model is all-in-one and can restore various real-world video degradation with a single diffusion model under the guidance of task prompts. (b) Our model can generate temporally consistent video frames with better preservation of original contents included in the input video.}
  \vspace{5mm}
  \label{fig:teaser}
\end{figure*}

In recent years, diffusion models have markedly advanced image generation \cite{LatentDiffusion} and video generation \cite{Text2Video-zero,tuneavideo,controlvideo}, becoming potential keystones in vision-based AI. Extensive research has demonstrated their ability to parse and encode diverse visual representations from massive text-to-image datasets, enriching downstream applications with strong real-world generalization for tasks such as image editing~\cite{instructpix2pix,nulltext} and classification~\cite{diffclass1}.
However, the random nature of diffusion models often disrupts the preservation of original image contents, presenting challenges for image restoration tasks. Some works have developed a diffusion model for each specific task \cite{diff_deblur,palette} by training it from scratch on a certain dataset, limiting its utility to a single task.
Alternatively, approaches like Denoising Diffusion Implicit Models Inversion (DDIM Inversion) \cite{ddim_inv,nulltext} and a content-preserving module with a task-wise plugin \cite{diffplugin} guide the diffusion process using input images for multi-target image editing and restoration. However, when these methods are directly applied to video restoration through image-by-image inference, they show insufficient temporal consistency.

To address the above issues, we propose the first diffusion-based all-in-one video restoration method, named Temporally-consistent Diffusion Model (TDM). TDM utilizes the capabilities of the pre-trained text-to-image Stable Diffusion (SD) model \cite{LatentDiffusion} alongside a fine-tuned ControlNet \cite{controlnet}. TDM is designed to offer a simple, efficient, and easy-to-extend framework for video restoration, focusing on improving training and inference strategies for a diffusion model without introducing complex modules.


For the training phase, we fine-tune a single-image-based ControlNet \cite{controlnet} on multiple tasks using text prompts referring to the task name, such as "remove the noise" for video denoising and "remove the rain" for video deraining. We refer to this as Task Prompt Guidance (TPG), which directs the diffusion process toward a specific restoration task. This approach leverages the strong zero-shot classification capability of the pre-trained SD model \cite{diffclass1}. This simple strategy enables robust all-in-one restoration on multiple tasks without additional computation time and parameters, as demonstrated in Fig.~\ref{fig:teaser}(a).
By focusing on fine-tuning ControlNet rather than training an entire diffusion model, we harness the generative power of the pre-trained SD model. This approach improves the quality of video restoration and enhances the robustness against real-world data. Furthermore, because ControlNet is fine-tuned using single-image inputs, our strategy eliminates the extreme memory requirements needed to process multiple video frames together during training, which is common in earlier methods \cite{text2video_svd,text2video_animetediff,text2video_modelscope} for ensuring temporal consistency. This allows our model to be trained on a single GPU using common single-image restoration datasets, simplifying data preparation and enhancing accessibility for future research.

In the inference phase, we aim to ensure content preservation and temporal consistency. Previous text-to-video studies \cite{Text2Video-zero,controlvideo} addressed temporal consistency by replacing the self-attention in U-Net with cross-frame attention during inference. However, the method~\cite{Text2Video-zero} struggles with large motions between video frames and the method~\cite{controlvideo} requires extensive memory and computational resources.
Also, both methods do not consider content preservation.
To address these issues, we propose a Sliding Window Cross-Frame Attention (SW-CFA) mechanism, combined with DDIM Inversion \cite{ddim_inv,nulltext}. This combination effectively tackles both content preservation and temporal consistency. Specifically, SW-CFA extends the concept of reference frames to a sliding window around the current frame, employing mean-based temporal smoothing in the attention calculation with minimal computational increase. This allows SW-CFA to handle large motions better than \cite{Text2Video-zero}, achieving effective zero-shot image-to-video adaptation by adjusting the attention mechanism during inference. As discussed in \cite{diffplugin}, adding random Gaussian noise to the input latent can reduce the fidelity of diffusion-generated images. Therefore, we introduce DDIM Inversion for better content preservation. The combination of SW-CFA and DDIM Inversion maintains a deterministic and close distribution of input noises across adjacent frames, resulting in a more coherent video output. As illustrated in Fig.~\ref{fig:teaser}(b), compared with a straightforward combination of SD and ControlNet, our proposed TDM with SW-CFA and DDIM Inversion generates more temporally consistent video frames.

In summary, we build the first all-in-one video restoration diffusion model in this work. Our main contributions are as follows:

\begin{itemize}
    \item Training strategies: We achieve efficient training by fine-tuning a single ControlNet using single-image inputs without requiring video inputs. The proposed TPG offers a simple way to achieve cross-task robustness.
    \item Inference strategies: We incorporate DDIM Inversion and introduce a novel SW-CFA mechanism for zero-shot video inference. This combination ensures accurate content preservation and robust temporal consistency in restored videos.
    \item Scalability: Our proposed TDM can be trained with a single GPU and is adaptable for video inference after being trained on single-image restoration datasets. This flexibility allows straightforward expansion to other video restoration tasks using only single-image datasets.
    \item Generalization performance: We conduct extensive experiments on five restoration tasks, demonstrating strong generalization performance of our TDM to real-world data over existing regression-based and diffusion-based methods.
\end{itemize}

\section{Methodology} \label{sec: method}

In this section, we first introduce the preliminary on latent diffusion models in Sec.~\ref{preliminary}. Then, we introduce the details of our method in Sec \ref{method:training} (Training) and Sec \ref{method:inference} (Inference).

\subsection{Preliminary}
\label{preliminary}

\textbf{Latent Diffusion Model }(LDM)~\cite{LatentDiffusion} significantly advances traditional diffusion models by operating in a latent space. LDM employs an encoding mechanism to transform an image $x$ into a compressed latent representation $z = E(x)$, thereby facilitating the learning of the latent codes' distribution, denoted as $z_0 \sim p_{\text{data}}(z_0)$ where $p_{\text{data}}$ denotes the latent distributions of training images, following the Denoising Diffusion Probabilistic Model~(DDPM) strategy introduced by \cite{ddpm}. First, LDM follows a forward phase that introduces Gaussian noise progressively over time steps $t$ to derive $z_t$:
\begin{equation}
    q(z_t|z_{t-1}) = \mathcal{N}(z_t; \sqrt{1 - \beta_t} z_{t-1}, \beta_t I),
\end{equation}
with the noise scale represented by $\{\beta_t\}_{t=1}^T$, the gaussian distribution by $\mathcal{N}$, an all-one tensor with the same shape as $z_t$ expressed by $I$, and the total diffusion steps denoted by $T$. Then, LDM follows a backward phase, where the model strives to reconstruct the preceding less noisy state $z_{t-1}$ as follows:
\begin{equation}
p_\theta(z_{t-1}|z_t) = \mathcal{N}(z_{t-1}; \mu_\theta(z_t, t, \tau), \Sigma_\theta(z_t, t, \tau)),
\end{equation}
where $\mu_\theta$ and $\Sigma_\theta$ are mean and variance of the current state characterized by learnable parameters $\theta$, and implemented with a noise prediction model $\epsilon_\theta$. To generate novel samples,  initialization starts with a Gaussian sample $z_T \sim \mathcal{N}(0, 1)$, followed by a DDIM backward process of $z_{t-1}$ for preceding time steps:
\begin{equation}
z_{t-1} = \sqrt{\alpha_{t-1}} \left( \frac{z_t - \sqrt{1-\alpha_t}\epsilon_\theta(z_t, t, \tau)}{\sqrt{\alpha_t}} \right) + \sqrt{1-\alpha_{t-1}} \cdot \epsilon_\theta(z_t, t, \tau),
\end{equation}
where the cumulative product $\alpha_t = \prod_{i=1}^{t}(1 - \beta_i)$ is used for simplicity to indicate the transition towards $z_0$ at step $t$. The most well-known latent diffusion model is SD~\cite{LatentDiffusion}, which exemplifies text-to-image LDMs trained on a vast corpus of image-text pairs, with  $\tau$ representing the text prompt. Moreover, it demonstrates its scalability across various other tasks~\cite{syncdreamer,Zero123,Maingold}.


\textbf{ControlNet}~\cite{controlnet} enhances the capabilities of SD for more precise text-to-image synthesis by incorporating additional inputs such as depth maps, body poses, and images. While retaining the U-Net architecture identical to that of SD, ControlNet is fine-tuned to accommodate specific conditional inputs, modifying the function from $\epsilon_\theta(z_t, t, \tau)$ to $\epsilon_\theta(z_t, t, c, \tau)$, where $c$ represents these additional conditional inputs.

\textbf{DDIM Inversion}~\cite{ddim_inv,ddim} is an inverse process of the DDIM forward process, based on the assumption that the ordinary differential equation can be reversed in the limit of small steps:
\begin{equation}
z_{t+1} = \sqrt{\alpha_{t+1}} \left( \frac{z_t - \sqrt{1-\alpha_t}\epsilon_\theta(z_t, t, \tau)}{\sqrt{\alpha_t}} \right) + \sqrt{1-\alpha_{t+1}} \cdot \epsilon_\theta(z_t, t, \tau).
\end{equation}
In other words, the diffusion process is performed in the reverse direction, that is $z_0$ to $z_T$ instead of $z_T$ to $z_0$, where $z_0$ in our case is set to be the encoding of the given input degraded image. Despite using ControlNet, we observe that employing noises generated through DDIM Inversion, rather than relying on random Gaussian noises, provides stable structural guidance that better preserves the contents of the input image.

\begin{figure*}[!th]
    \centering
\includegraphics[width=1.0\linewidth]{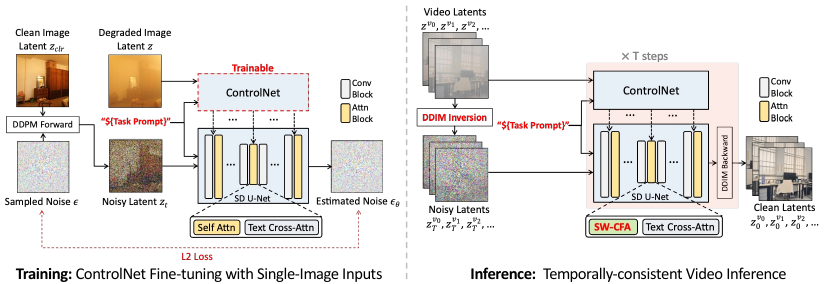}
    \caption{Overall architecture of our proposed temporally-consistent diffusion model (TDM).}
\label{fig:MainArchitecture}
\end{figure*}

\subsection{Training: Task Prompt Guided ControlNet Fine-Tuning with Single-Image Inputs}
\label{method:training}

The pre-trained SD model excels at generating high-quality images without degradation, a feature we strive to maintain, particularly in the face of severe degradations such as sensor noise, low resolution, and haze. To this end, as shown in Fig.~\ref{fig:MainArchitecture}, we leverage ControlNet~\cite{controlnet} which is a copy of the SD U-Net architecture. We fine-tune the ControlNet model initialized as the official tile resample model from ControlNet for SD version 1.5 because it is a solid foundation for image-to-image tasks. This ControlNet is then fine-tuned on our datasets, which include pairs of degraded images and their clean ground-truth images, focusing on image restoration.

\textbf{Task Prompt Guidance (TPG).} 
In our experiments, we consider five video restoration tasks: dehazing, deraining, denoising, MP4 compression artifact removal, and super-resolution. Leveraging the capacity of SD to generate diverse image styles through different text prompts, we recognize their potential to address a variety of tasks within a single model framework. To enhance this versatility, we introduce TPG, a method that employs specific task descriptions as input text prompts  ($\tau$) for both training and inference, as shown in Fig.~\ref{fig:teaser} and Fig.~\ref{fig:MainArchitecture}. Unlike the approach in \cite{diffplugin}, which relies on separate plugin modules for each task and a classifier to select the appropriate plugin based on the prompt, we utilize the proven proficiency of the pre-trained SD model to interpret and classify directly from text prompts as noted in studies~\cite{diffclass1,diffclass2}.

\textbf{Single-Image Inputs During Training.} 
As illustrated in Fig.~\ref{fig:MainArchitecture}, our proposed method requires only single-image inputs for video restoration tasks during the fine-tuning of the ControlNet. Drawing inspiration from previous works~\cite{Text2Video-zero,controlvideo}, we achieve significant temporal consistency through training-free operations at the inference stage, thereby circumventing the high memory demands of processing multiple video frames simultaneously during training, which is a common requirement in earlier approaches~\cite{text2video_svd,text2video_animetediff,text2video_modelscope} for achieving temporal consistency. This strategy allows our model to be trained efficiently on a single GPU using single-image restoration datasets, significantly reducing data preparation complexity and making it more accessible for future research. This approach paves the way for future research to easily append novel tasks with training on corresponding single-image datasets.

\begin{figure}[!th]
    \centering
\includegraphics[width=1.0\linewidth]{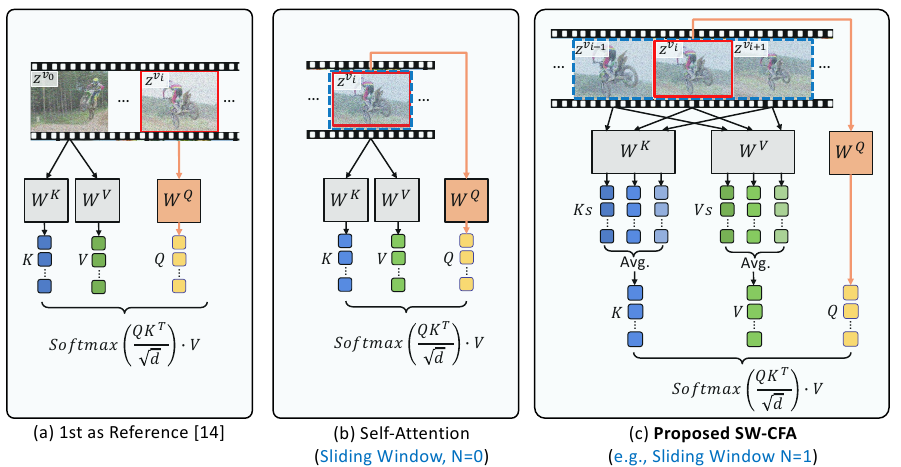}
    \vspace{-6mm}
    \caption{Proposed SW-CFA compared with exisiting cross-frame attention.}
\label{fig:LSCVA}
\end{figure}

\subsection{Inference: Training-Free Content-Preserved Temporal Consistency for Larger Motion}
\label{method:inference}

As mentioned earlier, we only use single-image inputs in the training phase, so temporal consistency is addressed through a training-free approach during the inference. Previous studies \cite{Text2Video-zero,controlvideo} achieve this by replacing self-attention layers in denoising U-Nets with cross-frame attention layers. LDM models utilize a U-Net architecture with downscaling and upscaling phases, enhanced with skip connections. The architecture includes 2D convolutional residual blocks and transformer blocks, each containing a self-attention layer, a cross-attention layer, and a feed-forward network. Self-attention formulates spatial correlations within the feature map, while cross-attention maps relationships between the feature map and external conditions like text prompts. For a video frame $v_i$ with latent representation $z^{v_i}$, as shown in Fig.~\ref{fig:LSCVA}(b), self-attention is defined as: $Attention(Q, K, V) = Softmax\left(\frac{QK^T}{\sqrt{d}}\right) \cdot V$, 
where 
\begin{equation}
    Q=W^Qz^{v_i}, K=W^Kz^{v_i}, V=W^Vz^{v_i},
\end{equation}
with $W^Q$, $W^K$, and $W^V$ being trainable matrices for query, key, and value projections, respectively, and $d$ is the dimension of key and query features.
Previous studies \cite{Text2Video-zero} reorganize self-attention into cross-frame attention using a single reference latent $z_{v_{ref}}$ as both the key ($K$) and value ($V$). Typically, this reference frame is set as the first frame ($v_0$) as shown in Fig.~\ref{fig:LSCVA}(a):
\begin{equation}
    Q=W^Qz^{v_i}, K=W^Kz^{v_0}, V=W^Vz^{v_0},
\end{equation}
This captures the spatial relationship between the current frame's query and the reference frame's key, maintaining the appearance, structure, and identities across frames, thus enhancing temporal consistency. However, using only $v_0$ as the reference limits motion range, as large motions may result in limited overlap between $v_0$ and subsequent frames. Additionally, the consistency between later consecutive frames is not explicitly addressed, further limiting overall temporal consistency even with small motions. 


\textbf{Sliding Window Cross-Frame Attention (SW-CFA)}. We propose a novel cross-frame attention mechanism named SW-CFA to accommodate a wider range of motions. As shown in Fig.~\ref{fig:LSCVA}(c), instead of relying on a single fixed reference frame for the key ($K$) and value ($V$), we extend the reference frames to include frames within a local window. Specifically, we average the keys and values inside the window, which includes both the preceding and succeeding $N$ frames. The formulation is as follows:
\begin{equation}
    Q=W^Qz^{v_i}, K = \frac{1}{2N + 1} \sum_{j=i-N}^{i+N} W^Kz^{v_j}, V = \frac{1}{2N + 1} \sum_{j=i-N}^{i+N} W^Vz^{v_j},
\end{equation}
which can be further simplified as
\begin{equation}
    Q=Q_i, K = \frac{1}{2N + 1} \sum_{j=i-N}^{i+N} K_j, V = \frac{1}{2N + 1} \sum_{j=i-N}^{i+N} V_j,
\end{equation}
and our SW-CFA is formulated as follows:
\begin{small}
\begin{equation}
    Attention(Q, K, V) = Softmax\left(Q_i\frac{1}{2N+1} \sum_{j=i-N}^{i+N} K_j^T \right) \left(\frac{1}{2N+1} \sum_{j=i-N}^{i+N} V_j\right).
\end{equation}    
\end{small}

By averaging $K_j$ and $V_j$ across a local window of frames, SW-CFA smooths over rapid variations, focusing on stable and consistent features. This is not merely filtering but a principled integration of temporal information, enhancing the model's ability to prioritize relevant information across temporal axis.

Mathematically, this averaging process reduces variance in the attention mechanism's input, effectively functioning as a temporal low-pass filter. This reduction allows the attention mechanism to produce outputs less sensitive to frame-to-frame fluctuations, increasing temporal consistency. Within the softmax function, averaging indirectly weights each nearest $N$ frames' $K$ and $V$ matrices according to their temporal proximity and similarity to the current frame $i$, leveraging the softmax function's property of amplifying significant signals while attenuating weaker ones.
Thus, SW-CFA captures both spatial dependencies within frames and temporal dependencies. This results in a robust attention mechanism that enhances temporal consistency through a simple yet effective modification of the previous cross-frame attention paradigm.


\textbf{Combination with DDIM Inversion}. 
Our proposed SW-CFA mechanism significantly enhances temporal consistency. However, random Gaussian noise added to input latents, as noted in \cite{diffplugin}, can reduce the fidelity of diffusion-generated images. To address this, we integrate DDIM Inversion \cite{ddim_inv,ddim} to provide stable input noise for each video frame's latents, serving as a solid structural guide, as shown in Fig.~\ref{fig:MainArchitecture}. DDIM Inversion ensures the noise added to each input frame's latents is deterministic, consistent, and derived from the latents themselves. Since these input latents originate from temporally coherent video frames, the noise introduced by DDIM Inversion also exhibits temporal coherence. This synergy eliminates the disruption caused by random Gaussian noise, enabling SW-CFA to more precisely capture temporal relationships. Therefore, combining DDIM Inversion not only improves content preservation but also further strengthens temporal consistency.

\section{Experiments}\label{sec:exp}

\subsection{Settings}\label{sec:ablation}

Here, we outline our experimental settings, including the datasets, implementation details, and evaluation metrics.

\textbf{Datasets.} We evaluated the proposed TDM for five video restoration tasks. To train the TDM and the compared methods, we utilized representative datasets for each task, dehazing: REVIDE~\cite{REVIDE}, deraining: NTURain-syn~\cite{semividdenoise_nturain}, denoising: DAVIS~\cite{davis}, MP4 compression artifacts removal: MFQEv2~\cite{mfqev2}, and super-resolution~(SR): REDS~\cite{reds}. Only REVIDE is a real-world dataset and the others are synthetic datasets. To avoid unbalanced training due to largely different amounts of training images for each task, we adjusted the number of images from each dataset to around 5,000-6,000 images. Then, we combined all of those images to construct the training dataset, resulting in a total of 27,843 training images. 
For testing, we used real-world benchmark datasets for each task, dehazing: REVIDE~\cite{REVIDE} (284 images), denoising: CRVD~\cite{rawviddenoise} (560 images), deraining: NTURain-real~\cite{semividdenoise_nturain} (658 images), MP4 compression artifacts removal: MFQEv2~\cite{mfqev2} (1,080 images), and SR: UDM10~\cite{udm10} (320 images). 

\textbf{Implementation.} 
During the training and the inference phase, we adjusted the size of each image by resizing it to ensure that the shorter side is 512 pixels. For training, we randomly cropped patches of 512x512 pixels from resized images. We utilized the AdamW optimizer with its default settings, including betas and weight decay. The training of our ControlNet was carried out with a constant learning rate of 1e-5 and a batch size of 4, on a single RTX 4090 GPU, and we incorporated gradient checkpoints for efficiency. The training process lasts for 25 epochs.
In the inference stage, we set the window radius $N$ for SW-CFA as 3 for all experiments. For the sampling technique, we initially applied DDIM Inversion~\cite{ddim_inv} with 10 timesteps, followed by DDIM backward sampling~\cite{ddim} using 32 timesteps. Our proposed TDM model can process a 15-frame video with resolutions of 512x896 pixels in under 30 seconds with a single RTX 4090 GPU while requiring 10GB of GPU memory.

\textbf{Metrics.} We followed \cite{diffplugin,LatentDiffusion} to employ widely adopted non-reference perceptual metrics, FID~\cite{fid} and KID~\cite{kid}, to evaluate our TDM, as the ground-truth images are not always available for real-world datasets. For easier view, the KID value is scaled by 100$\times$. For temporal consistency evaluation, we followed \cite{controlvideo} to estimate (i) Frame consistency (FC): the average cosine similarity between all pairs of consecutive frames, and (ii) Warping Error (WE): the average mean squared error of consecutive frames after aligning the next frame to the current one using estimated optical flow. For easier view, FC is scaled by 10$\times$ and WE is scaled by 1000$\times$.

\subsection{Comparison with State-of-the-Art Methods}\label{sec:compare}

We compared our TDM with six state-of-the-art methods for image/video restoration: AirNet~\cite{airnet}, PromptIR~\cite{promptir}, VRT~\cite{vrt}, RVRT~\cite{rvrt}, WeatherDiff~\cite{weatherdiffusion}, and InstructP2P~\cite{instructpix2pix}. 
As the category of each method, AirNet, PromptIR, VRT, and RVRT are regression-based methods, such as based on CNNs or Transformers, while 
WeatherDiff, InstructP2P, and our TDM are diffusion-based methods. As the input for the inference, AirNet, PromptIR, WeatherDiff, InstructP2P use a single-image input, while VRT, RVRT, and our TDM use a video input.
We trained all methods using the same training dataset and tested them for real-world testing datasets, as explained in the previous subsection.

\begin{figure*}[!t]
    \centering
\includegraphics[width=1.0\linewidth]{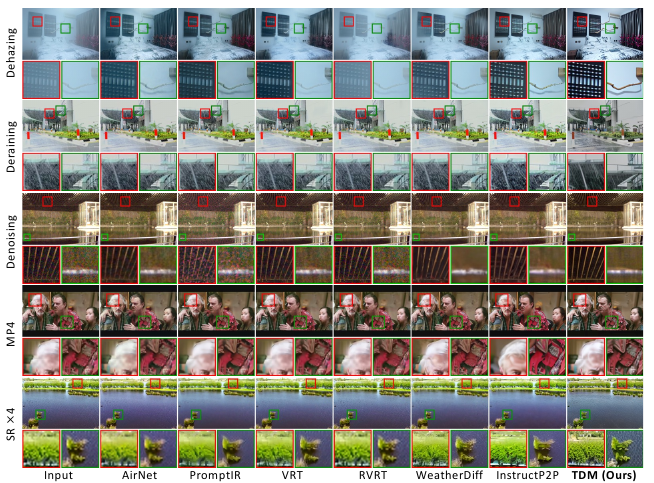}
    \vspace{-7mm}
    \caption{Qualitative comparison with state-of-the-art methods.}
    \vspace{-2mm}
\label{fig:compare}
\end{figure*}

\textbf{Quantitative results.} 
We provide a quantitative comparison in Table~\ref{tb:quantitative}. It demonstrates the superiority of diffusion-based methods in producing higher-quality images, as evidenced by better FID and KID scores compared to regression-based methods. Compared with other regression- and diffusion-based methods, our proposed TDM achieves the best results on average. Although InstructP2P also generates high-quality images with low FID and KID, it tends to alter the original contents, as seen in Fig.~\ref{fig:compare}.
Compared to the regression-based methods including AirNet and PromptIR, which are designed for generalization across different tasks, our TDM still demonstrates robust cross-task performance with the guidance of the proposed TPG. Furthermore, while video restoration methods such as VRT and RVRT are designed to utilize video temporal information effectively during the training phase, they struggle with multi-task handling. In contrast, our TDM consistently delivers state-of-the-art performance in video restoration tasks, even though it is trained on single-image inputs.

\begin{table*}[t!]
    \centering
    \caption{Quantitative comparison with state-of-the-art methods (Red: rank 1st; Blue: rank 2nd).}
    \aboverulesep=0ex
    \belowrulesep=0ex
    \label{tb:quantitative}
    \renewcommand{\arraystretch}{1.3}
    \renewcommand\tabcolsep{1pt}
    \vspace{-2mm}
    \scriptsize
    \begin{tabular}{c|c|cc|cc|cc|cc|cc|cc}
        \toprule
        \multirow{2}{*}{Method Types} & \multicolumn{1}{c|}{\multirow{2}{*}{Methods}} & \multicolumn{2}{c|}{Dehazing} & \multicolumn{2}{c|}{Deraining} & \multicolumn{2}{c|}{Denoising} & \multicolumn{2}{c|}{MP4} & \multicolumn{2}{c|}{SR $\times$4} & \multicolumn{2}{c}{Average} \\
        \cmidrule{3-14}
                                    & & FID$\downarrow$ & KID$\downarrow$ & FID$\downarrow$ & KID$\downarrow$ & FID$\downarrow$ & KID$\downarrow$ & FID$\downarrow$ & KID$\downarrow$ & FID$\downarrow$ & KID$\downarrow$ & FID$\downarrow$ & KID$\downarrow$ \\
        \midrule
        \multirow{2}{*}{Single-image regression} 
        & AirNet~\cite{airnet}        & 83.34 & 7.77 & 90.86 & 4.24 & 80.63 & 4.41 & 104.48 & \textcolor{red}{\textbf{6.56}} & 88.09 & 3.07 & 89.47 & 5.21 \\
        & PromptIR~\cite{promptir}    & 75.21 & \textcolor{red}{\textbf{5.73}} & 88.47 & 5.28 & 82.11 & 4.95 & 102.73 & 6.68 & 89.37 & 3.12 & 87.57 & 5.15 \\
        \midrule
        \multirow{2}{*}{Video regression} 
        & VRT~\cite{vrt}           & 79.88 & 7.03 & 88.36 & 4.69 & 81.94 & 4.88 & 107.00 & \textcolor{red}{\textbf{6.56}} & 88.97 & 2.71 & 89.23 & 5.17    \\
        & RVRT~\cite{rvrt}            & 79.27 & 6.55 & 94.13 & 5.44 & 83.59 & 4.57 & 107.37 & 6.55 & 88.51 & 2.93 & 90.57 & 5.20    \\
        \midrule
        \multirow{2}{*}{Single-image diffusion} 
        & WeatherDiff~\cite{weatherdiffusion} & 74.42 & 6.22 & 88.20 & 4.85 & 80.11 & 4.18 & 102.24 & 6.67 & \textcolor{blue}{\textbf{88.07}} & 2.77 & 86.28 & 4.93      \\
        & InstructP2P~\cite{instructpix2pix} & \textcolor{blue}{\textbf{74.02}} & \textcolor{blue}{\textbf{6.01}} & \textcolor{blue}{\textbf{81.51}} & \textcolor{red}{\textbf{3.36}} & \textcolor{blue}{\textbf{79.71}} & \textcolor{red}{\textbf{3.90}} & \textcolor{blue}{\textbf{101.62}} & 7.70 & 88.21 & \textcolor{blue}{\textbf{2.62}} & \textcolor{blue}{\textbf{85.01}} & \textcolor{blue}{\textbf{4.71}}    \\
        \midrule
        \multirow{1}{*}{Video diffusion} 
        & \textbf{TDM~(Ours)}         & \textcolor{red}{\textbf{73.68}} & 6.42 & \textcolor{red}{\textbf{81.27}} & \textcolor{blue}{\textbf{3.79}} & \textcolor{red}{\textbf{78.63}} & \textcolor{blue}{\textbf{4.09}} & \textcolor{red}{\textbf{100.91}} & \textcolor{blue}{\textbf{6.59}} & \textcolor{red}{\textbf{88.04}} & \textcolor{red}{\textbf{2.55}} & \textcolor{red}{\textbf{84.50}} & \textcolor{red}{\textbf{4.68}} \\
        \bottomrule
    \end{tabular}
\end{table*}

\textbf{Qualitative results.} Figure~\ref{fig:compare} illustrates the robust performance of our TDM across five challenging real-world video restoration tasks. 
In the denoising task (1st row), while most regression-based methods trained on synthesized Gaussian noise struggle to recognize real-world noise patterns, all diffusion-based methods including TDM successfully eliminate the real-world noise. However, TDM distinctly outperforms WeatherDiff and InstructP2P, which both leave artifacts post-denoising.
In the dehazing task (2nd row), TDM excels by clearing heavy fog to reveal the sharpest and the most haze-free images. 
For the MP4 compression artifact removal task (3rd row), TDM outperforms regression-based methods in detail restoration. Compared with diffusion-based InstructP2P, which inaccurately alters hairstyles, our TDM can restore sharp details and maintain consistency with the original input. 
In the SR task (4th row), TDM achieves the clearest detail enhancement, whereas other methods produce comparatively blurred results. 
Finally, in the challenging heavy rain scenario (5th row), while other models fail to detect and remove the rain, TDM significantly reduces rain artifacts, demonstrating its robustness in severe weather conditions.
Overall, our TDM consistently delivers superior restoration quality with remarkable detail preservation and consistency across different video restoration challenges.

\begin{figure}[!t]
    \centering
\includegraphics[width=0.77\linewidth]{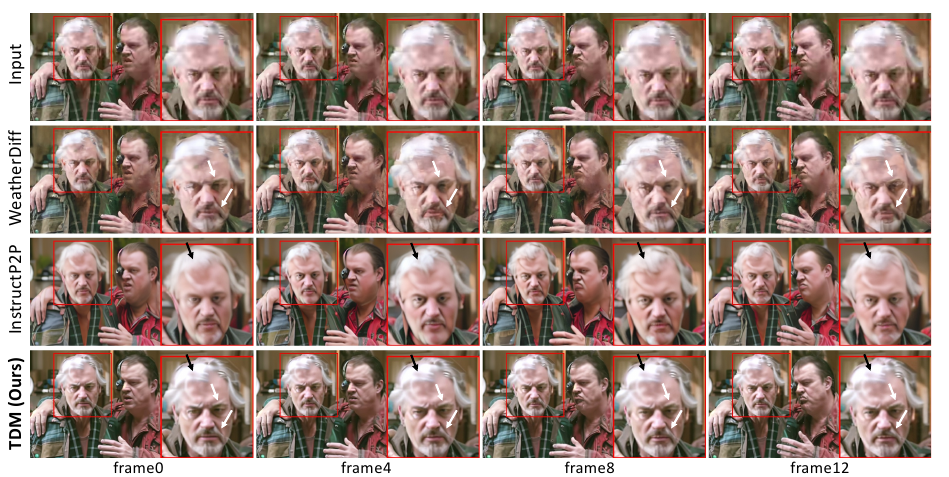}
\vspace{-3mm}
    \caption{Consistency comparison (MP4) with other diffusion-based methods.}
\vspace{-1mm}
\label{fig:consistency_compare}
\end{figure}

\begin{figure}[!t]
    \centering
\includegraphics[width=0.77\linewidth]{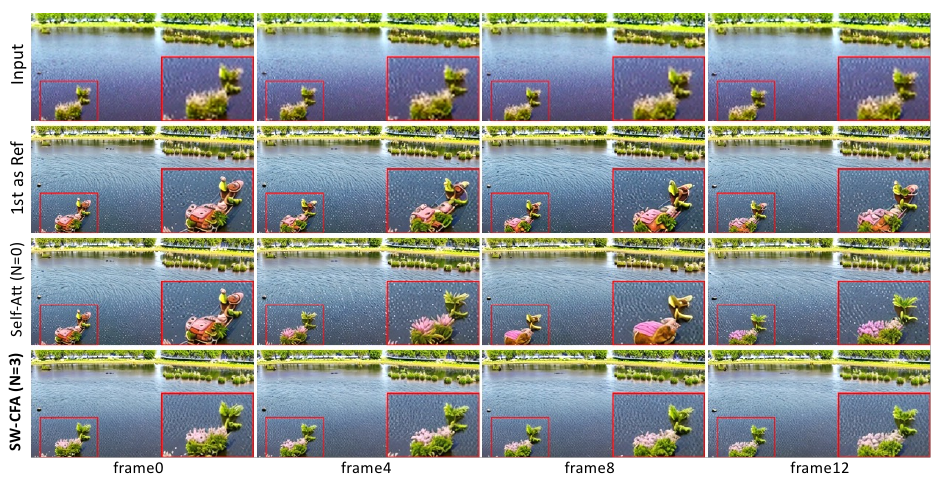}
\vspace{-3mm}
    \caption{Consistency comparison (SR$\times$4) with other zero-shot cross-frame attention.}
\vspace{-5mm}
\label{fig:consistency_compare_att}
\end{figure}

\textbf{Consistency Evaluation.} 
Although diffusion-based methods typically yield images of high visual quality, their inherent randomness often leads to poor temporal consistency in video processing. In Table~\ref{tb:consistency_compare}, we compare our TDM against other diffusion-based techniques using two temporal consistency metrics: WE and FC. The results show that TDM consistently outperforms the others in maintaining temporal consistency according to both metrics.
Figure~\ref{fig:consistency_compare} provides a visual comparison that supports these results. While the methods WeatherDiff and InstructP2P can remove the noise, they exhibit notable inconsistencies between the frames. WeatherDiff (2nd row) exhibits heavy changes in the appearance of building frames in the background. For InstructP2P (3rd row), despite using image-based classifier-free guidance, it still displays fluctuation across frames. In contrast, TDM, utilizing DDIM inversion to eliminate randomness and SW-CFA to bolster temporal stability, maintains consistent features across all frames, leading to stable denoising results. This robustness underscores TDM's superior temporal consistency compared to other diffusion-based methods. 
Also, in Table~\ref{tb:consistency_zeroshot} and Fig.~\ref{fig:consistency_compare_att}, we evaluate our proposed SW-CFA against the cross-frame attention from Text2Video-zero~\cite{Text2Video-zero} (referred to as "1st as Ref") and the standard self-attention ($N$=0). While "1st as Ref" enhances consistency by using the first frame as a reference, it overlooks the consistency of consecutive frames. In contrast, our SW-CFA, which averages key-value pairs within a sliding window, achieves more substantial consistency improvements. It demonstrates a superior ability to maintain uniformity across frames, effectively handling larger motions and providing robust temporal stability.

\begin{table*}[t!]
\caption{Image quality and consistency comparison with other diffusion-based methods.}
\vspace{-5mm}
\label{tb:consistency_compare}
\scriptsize
\begin{center}
\renewcommand\tabcolsep{1.5pt}
\renewcommand\arraystretch{1.4}
\begin{tabular}{c|c c c | c c c | c c c | c c c | c c c } \hline
     \multirow{2}{*}{Methods} & \multicolumn{3}{c|}{Dehazing} & \multicolumn{3}{c|}{Deraining} & \multicolumn{3}{c|}{Denoising} & \multicolumn{3}{c|}{MP4} & \multicolumn{3}{c}{SR $\times$4} \\ 
     \cline{2-16} 
     & FID$\downarrow$ & FC$\uparrow$ & WE$\downarrow$ & FID$\downarrow$ & FC$\uparrow$ & WE$\downarrow$ & FID$\downarrow$ & FC$\uparrow$ & WE$\downarrow$ & FID$\downarrow$ & FC$\uparrow$ & WE$\downarrow$ & FID$\downarrow$ & FC$\uparrow$ & WE$\downarrow$  \\ 
     \hline
     WeatherDiff~\cite{weatherdiffusion} & 74.42 & 9.759 & 8.537 & 88.20 & 9.682 & 3.652 & 80.11 & 9.439 & 2.182 & 102.24 & 9.908 & 1.282 & 88.07 & \textbf{9.682} & 5.751 \\ 
     InstructP2P~\cite{instructpix2pix} & 74.02 & 9.663 & 6.994 & 81.51 & 9.676 & 4.339 & 79.71 & \textbf{9.436} & 2.628 & 101.62 & 9.877 & 2.126 & 88.21 & 9.628 & 5.929 \\ 
     \textbf{TDM~(Ours)} & \textbf{73.68} & \textbf{9.849} & \textbf{5.464} & \textbf{81.27} & \textbf{9.700} & \textbf{3.415} & \textbf{78.63} & 9.296 & \textbf{2.008} & \textbf{100.91} & \textbf{9.921} & \textbf{1.138} & \textbf{88.04} & 9.651 & \textbf{5.741} \\
     \hline
\end{tabular}
\vspace{-5mm}
\end{center}
\end{table*}

\begin{table*}[!t]
\caption{Image quality and consistency comparison with state-of-the-art zero-shot cross-frame attention (by replacing the self-attention with according cross-frame attention during inference).}
\vspace{-2mm}
\label{tb:consistency_zeroshot}
\begin{center}
\scriptsize
\renewcommand\tabcolsep{1.7pt}
\renewcommand\arraystretch{1.4}
\begin{tabular}{c|c c c | c c c | c c c | c c c | c c c } \hline
     \multirow{2}{*}{Methods} & \multicolumn{3}{c|}{Dehazing} & \multicolumn{3}{c|}{Deraining} & \multicolumn{3}{c|}{Denoising} & \multicolumn{3}{c|}{MP4} & \multicolumn{3}{c}{SR $\times$4} \\ 
     \cline{2-16} 
     & FID$\downarrow$ & FC$\uparrow$ & WE$\downarrow$ & FID$\downarrow$ & FC$\uparrow$ & WE$\downarrow$ & FID$\downarrow$ & FC$\uparrow$ & WE$\downarrow$ & FID$\downarrow$ & FC$\uparrow$ & WE$\downarrow$ & FID$\downarrow$ & FC$\uparrow$ & WE$\downarrow$ \\ 
     \hline
     1st as Ref.~\cite{Text2Video-zero} & 74.37 & 9.712 & 6.432 & 82.39 & 9.693 & 3.526 & 80.12 & 9.229 & 2.219 & \textbf{99.63} & 9.901 & 1.387 & 88.62 & 9.622 & 6.350 \\ 
     Self-Attn. ($N$=0) & \textbf{73.36} & 9.703 & 6.801 & 83.06 & 9.690 & 3.596 & 80.17 & 9.213 & 2.368 & 100.56 & 9.896 & 1.459 & 88.30 & 9.618 & 6.769 \\ 
     \textbf{SW-CFA ($N$=3)} & 73.68 & \textbf{9.849} & \textbf{5.464} & \textbf{81.27} & \textbf{9.700} & \textbf{3.415} & \textbf{78.63} & \textbf{9.296} & \textbf{2.008} & 100.91 & \textbf{9.921} & \textbf{1.138} & \textbf{88.04} & \textbf{9.651} & \textbf{5.741}   \\ 
     \hline
\end{tabular}
    \vspace{-5mm}
\end{center}
\end{table*}

\subsection{Ablation Study}\label{sec:ablation}

In Table~\ref{tb:ablation}, we present an ablation study to assess the contribution of each component of our proposed TDM. We test three configurations: T+I, which omits the SW-CFA; T+S, which excludes DDIM Inversion (labeled as Inv.); and I+S, which lacks the TGP. The results show that removing either SW-CFA or DDIM Inversion diminishes temporal consistency, while their combination significantly enhances it. This indicates that DDIM Inversion, by stabilizing the input noise across frames, substantially supports SW-CFA in maintaining consistency.
Furthermore, the setup without TPG achieves comparable consistency scores, but there is a noticeable degradation in image quality, as evidenced by the increased FID scores. This underscores the critical role of TPG in boosting cross-task generalization and improving overall image quality in various restoration tasks.
\begin{table*}[!t]
\caption{Ablation study.}
\vspace{-5mm}
\label{tb:ablation}
\begin{center}
\scriptsize
\renewcommand\tabcolsep{0.5pt}
\renewcommand\arraystretch{1.4}
\begin{tabular}{c|c|c|c|c|c|c|c|c|c|c|c|c|c|c|c|c|c|c} \hline
     \multirow{2}{*}{Methods} & \multicolumn{3}{c|}{Proposals} & \multicolumn{3}{c|}{Dehazing} & \multicolumn{3}{c|}{Deraining} & \multicolumn{3}{c|}{Denoising} & \multicolumn{3}{c|}{MP4} & \multicolumn{3}{c}{SR $\times$4} \\ 
     \cline{2-19} 
     & TPG & Inv. & SW-CFA & FID$\downarrow$ & FC$\uparrow$ & WE$\downarrow$ & FID$\downarrow$ & FC$\uparrow$ & WE$\downarrow$ & FID$\downarrow$ & FC$\uparrow$ & WE$\downarrow$ & FID$\downarrow$ & FC$\uparrow$ & WE$\downarrow$ & FID$\downarrow$ & FC$\uparrow$ & WE$\downarrow$ \\ 
     \hline
     T+I & \checkmark & \checkmark &  &  \textbf{73.36} & 9.703 & 6.801 & 83.06 & 9.690 & 3.596 & 80.17 & 9.213 & 2.368 & \textbf{100.56} & 9.896 & 1.459 & 88.30 & 9.618 & 6.769 \\
     T+S & \checkmark &  & \checkmark & 73.45 & 9.738 & 7.121 & 82.79 & 9.673 & 4.125 & 80.07 & 9.207 & 3.792 & 101.47 & 9.866 & 1.841 & 88.35 & 9.552 & 8.418 \\
     I+S &  & \checkmark & \checkmark & 77.56 & 9.838 & 5.653 & 87.28 & 9.698 & 3.437 & 82.68 & 9.273 & 2.296 & 102.44 & 9.909 & 1.190 & 91.02 & 9.630 & 5.874 \\
     \textbf{Ours} & \checkmark & \checkmark & \checkmark & 73.68 & \textbf{9.849} & \textbf{5.464} & \textbf{81.27} & \textbf{9.700} & \textbf{3.415} & \textbf{78.63} & \textbf{9.296} & \textbf{2.008} & 100.91 & 9\textbf{.921} & \textbf{1.138} & \textbf{88.04} & \textbf{9.651} & \textbf{5.741} \\ 
     \hline
\end{tabular}
\vspace{-7mm}
\end{center}
\end{table*}

\section{Conclusion}\label{sec:conclusion}

In conclusion, we introduced the Temporally-consistent Diffusion Model (TDM) for all-in-one video restoration, utilizing a pre-trained Stable Diffusion model and fine-tuned ControlNet. Our method handles various video degradations with a single model, using Task Prompt Guidance (TPG) for training and combining DDIM Inversion with Sliding Window Cross-Frame Attention (SW-CFA) for enhanced temporally-consistent video inference.
Experiments across five tasks exhibits proposed TDM's superior generalization over existing methods, setting a new standard for video restoration. However, our method still falls short of regression-based methods in temporal consistency. Future work will focus on addressing this gap to further improve effectiveness.

\bibliographystyle{splncs04}
\bibliography{ref}
\end{document}